\documentclass[sigchi]{acmart}
\usepackage{nopageno}

\usepackage{booktabs} 
\usepackage{paralist} 
\usepackage{enumitem}
\setlist{nolistsep,leftmargin=*}

\usepackage{microtype}
\makeatletter
\renewcommand\section{\@startsection{section}{1}{\z@}%
  {-.45\baselineskip \@plus -2\p@ \@minus -.2\p@}%
  {.2\baselineskip}%
  {\@secfont}}
\renewcommand\subsection{\@startsection{subsection}{2}{\z@}%
  {-.4\baselineskip \@plus -2\p@ \@minus -.2\p@}%
  {.2\baselineskip}%
  {\@subsecfont}}
\renewcommand\subsubsection{\@startsection{subsubsection}{3}{10pt}%
  {-.4\baselineskip \@plus -2\p@ \@minus -.2\p@}%
  {-3.5\p@}%
  {\@subsubsecfont\@adddotafter}}
\makeatother
\usepackage[normalem]{ulem} 
\usepackage{xcolor}
\usepackage{multirow}
\usepackage{multicol}

\newcommand{\qianwen}[1]{#1}

\usepackage{marginnote}

\newcommand{\eg}{\textit{e.g.}}
\newcommand{\ie}{\textit{i.e.}}

\usepackage{setspace}
\usepackage{array}
\newcommand\Mark[1]{\textsuperscript#1}

\usepackage{balance}

\acmDOI{10.475/123_4}

\acmISBN{123-4567-24-567/08/06}

\copyrightyear{2019}
\acmYear{2019}
\setcopyright{acmlicensed}
\acmConference[CHI 2019]{CHI Conference on Human Factors in Computing Systems Proceedings}{May 4--9, 2019}{Glasgow, Scotland UK} 
\acmBooktitle{CHI Conference on Human Factors in Computing Systems Proceedings (CHI 2019), May 4--9, 2019, Glasgow, Scotland UK} \acmPrice{15.00}
\acmDOI{10.1145/3290605.3300911} \acmISBN{978-1-4503-5970-2/19/05}

\begin{document}
\title{
ATMSeer: Increasing Transparency and Controllability in Automated Machine Learning
}
 \settopmatter{authorsperrow=1}

 \author{Qianwen Wang\Mark{1}, Yao Ming\Mark{1}, Zhihua Jin\Mark{3}, Qiaomu Shen\Mark{1}, \\ Dongyu Liu\Mark{1}, Micah J. Smith\Mark{2}, \mbox{Kalyan Veeramachaneni}\Mark{2}, Huamin Qu\Mark{1}}
 \affiliation{%
  \begin{tabular}{*{3}{>{\centering}p{.25\textwidth}}}
   \Mark{1}Hong Kong University of \\Science and Technology & \Mark{2}Massachusetts Institute of Technology & \Mark{3}Zhejiang University \tabularnewline
   \{qwangbb, yao.ming, qshen, dliuae\}@ust.hk, huamin@cse.ust.hk &
   \{micahs, kalyanv\}@mit.edu & 
   jnzhihuoo1@gmail.com
  \end{tabular}
 }
 \email{}

\renewcommand{\shortauthors}{Q. Wang et al.}
\renewcommand{\shorttitle}{ATMSeer}

\fancyhead{}

\begin{abstract}
\qianwen{
To relieve the pain of manually selecting machine learning algorithms and tuning hyperparameters, automated machine learning (AutoML) methods have been developed to automatically search for good models.
Due to the huge model search space, it is impossible to try all models. Users tend to distrust automatic results and increase the search budget as much as they can, thereby undermining the efficiency of AutoML.
To address these issues, we design and implement ATMSeer, an interactive visualization tool that supports users in refining the search space of AutoML and analyzing the results.
To guide the design of ATMSeer, we derive a workflow of using AutoML based on interviews with machine learning experts.
A multi-granularity visualization is proposed to enable users to monitor the AutoML process, analyze the searched models, and refine the search space in real time.
We demonstrate the utility and usability of ATMSeer through two case studies, expert interviews, and a user study with 13 end users.
}

\end{abstract}

%
%




\maketitle

\begin{spacing}{1}

\section{Introduction}

To ease the difficulty of developing machine learning (ML) models, \qianwen{automated machine learning (AutoML) methods have been proposed~\cite{real2017large, li2017hyperband, zoph2016neural}.}
Instead of searching algorithms and tuning hyperparameters manually, AutoML automatically iterates through various machine learning algorithms and optimizes hyperparameters in a predefined search space (\ie, a set of feasible machine learning models). 
AutoML has received considerable research attention and gained widespread popularity.
A plethora of systems for AutoML, such as ATM~\cite{swearingen2017atm}, SigOpt~\cite{sigopt}, and Google Cloud AutoML~\cite{cloudAutoML} have been developed in recent years.

Unfortunately, these AutoML systems usually work as black boxes.
Due to the lack of transparency in AutoML (\eg, \textit{what models have been searched?}), 
users tend to question the automatic results. 
\textit{
Did the AutoML sufficiently explore the search space? 
Did the AutoML run long enough? 
Did the AutoML miss some suitable models?
}
Concerns like these may make users reluctant to apply the results of AutoML in critical applications~\cite{luhmann2018trust}, such as disease diagnosis and stock market prediction.
Meanwhile, when AutoML returns unsatisfying results, users are unable to reason and thus improve the results.
They can only increase the computational budget (\eg, running time) as much as possible, which undermines the efficiency of AutoML.

\qianwen{
These issues can be alleviated by involving end users in AutoML, enabling them to reason the AutoML results and to modify the AutoML configurations.
However, two challenges need to be addressed.
First, it can be difficult for users to analyze AutoML results.
An AutoML process generates a series of (usually a few hundred) models selected based on a specific search strategy.
These models have different algorithms, hyperparameter configurations, and performance scores.
It is non-trivial to organize and present this data in an intuitive way so that users can easily understand and analyze it.
Second, it can be challenging for users to modify the search space of an AutoML process.
AutoML can return unsatisfying models due to various reasons, such as insufficient budget, large search space, and limitations of AutoML algorithms ~\cite{swearingen2017atm, real2017large, AutoMLCompare}.
At the same time, the search space usually has a complicated hierarchical structure.
Effective interactions are required to help users modify an AutoML process by combining their observation of the current process with their prior knowledge.
}


In this paper, we present ATMSeer~\footnote{\url{https://github.com/HDI-Project/ATMSeer}}, an interactive visualization tool that helps users analyze the searched models and refine the search space.
\qianwen{Instead of opening the black box of AutoML and explaining the search decisions, }
ATMSeer offers a visual summary of the searched models to increase the \textbf{transparency} of AutoML.
Users are allowed to explore the models searched by an AutoML process at three levels of detail (\ie, the algorithm level, the hyperpartition level, and the hyperparameter level) \qianwen{
based on the breadth (\eg, has it searched all machine learning algorithms) and the depth (\eg, has it extensively searched algorithms that can lead to good performance)
}.
Meanwhile, ATMSeer enables users to interactively modify the search space in real time to increase the \textbf{controllability} of AutoML.
Through the visual summary of the searched models from three levels, users are able to understand the behavior of different models, which helps them \qianwen{propose alternative models and modify the search space}.
An in-situ search space configuration is embedded in the three-level visualization to facilitate the switch between analysis of the results and modification of the search space. 

In this work, we integrate ATMSeer with the ATM AutoML framework proposed by ~\citeauthor{swearingen2017atm}~\cite{swearingen2017atm}.
However, ATMSeer is not algorithm specific and can integrate with a variety of AutoML frameworks.

The main contributions of this paper are as follows:
\begin{itemize}
    \item A summary of the workflow for using AutoML tools and the requirements for analyzing an automated model search process.
    \item An interactive visualization tool that enables users to monitor, analyze, and refine an AutoML process.
    \item An evaluation of ATMSeer through two case studies, interviews with two AutoML experts, and a user study with 13 end users.
\end{itemize}

\section{Related Work}
\subsection{Choosing Machine Learning Models}
There is no one machine learning model that works the best for every problem~\cite{wolpert1996lack, fernandez2014we}.
To achieve high performance for a particular problem, 
users typically choose models based on their understanding of the algorithms, 
their observation of the data, 
and a time-consuming trial-and-error process.

Many efforts have been made to provide guidance for choosing models.
On the one hand, some research provides theoretical guidance by summarizing the pros and cons of different machine learning algorithms~\cite{box2005statistics, kotsiantis2007supervised}.
For example, \citeauthor{kotsiantis2007supervised}~\cite{kotsiantis2007supervised} conclude that support vector machines have a high tolerance for irrelevant features but require a large sample size. 
On the other hand, experiments on a large number of datasets also provide empirical guidance for choosing models~\cite{fernandez2014we, auto-sklearn, caruana2006empirical, ahmed2010empirical}.
For example, by evaluating 179 classifiers on 121 datasets,
\citeauthor{fernandez2014we}~\cite{fernandez2014we} find that random forests are most likely to be a good classifier, followed by support vector machines and neural networks.  

While these work provides useful guidance, they fail to provide detailed instruction for a particular problem (\eg, the exact model for a dataset).
ATMSeer aims to provide guidance to solve particular problems.
Given a dataset, ATMSeer automatically tries different models and allow users to easily observe and analyze these models through an interactive visualization.


\subsection{Visualizing Automated Machine Learning}
\qianwen{
Visualization has long been used to facilitate human interaction in the model tuning process~\cite{mcgregor2015facilitating, marks1997design}.
Recently, efforts have been taken to visualize automated machine learning.
}

For example, MLJar~\cite{MLJar} enables users to easily define a search space and analyze searched models with no coding required.
Google Vizer ~\cite{Golovin2017vizier} provides parallel coordinates to support the analysis of searched models.
For one algorithm, users can observe the range of each hyperparameters, the correlation between hyperparameters, and the relationship between performance and hyperparameters.
SigOpt~\cite{sigopt} provides an interface that enables users to join in the optimization loop of a model.
Users repeatedly observe suggested hyperparameter values, experiment with these values with their own model, analyze the experiment results, and finally report results back to SigOpt.

However, these works only support the analysis of one type of model (\eg, neural networks) at a time.
In contrast, ATMSeer supports the analysis of machine learning models generated with various algorithms (14 machine learning algorithms are supported in ATMSeer). 
Moreover, we provide a multi-granularity visualization of searched models to facilitate the analysis of the AutoML process.

\begin{figure*}[!htb]
\centering 
\vspace{-5mm}
\includegraphics[width=0.75\linewidth]{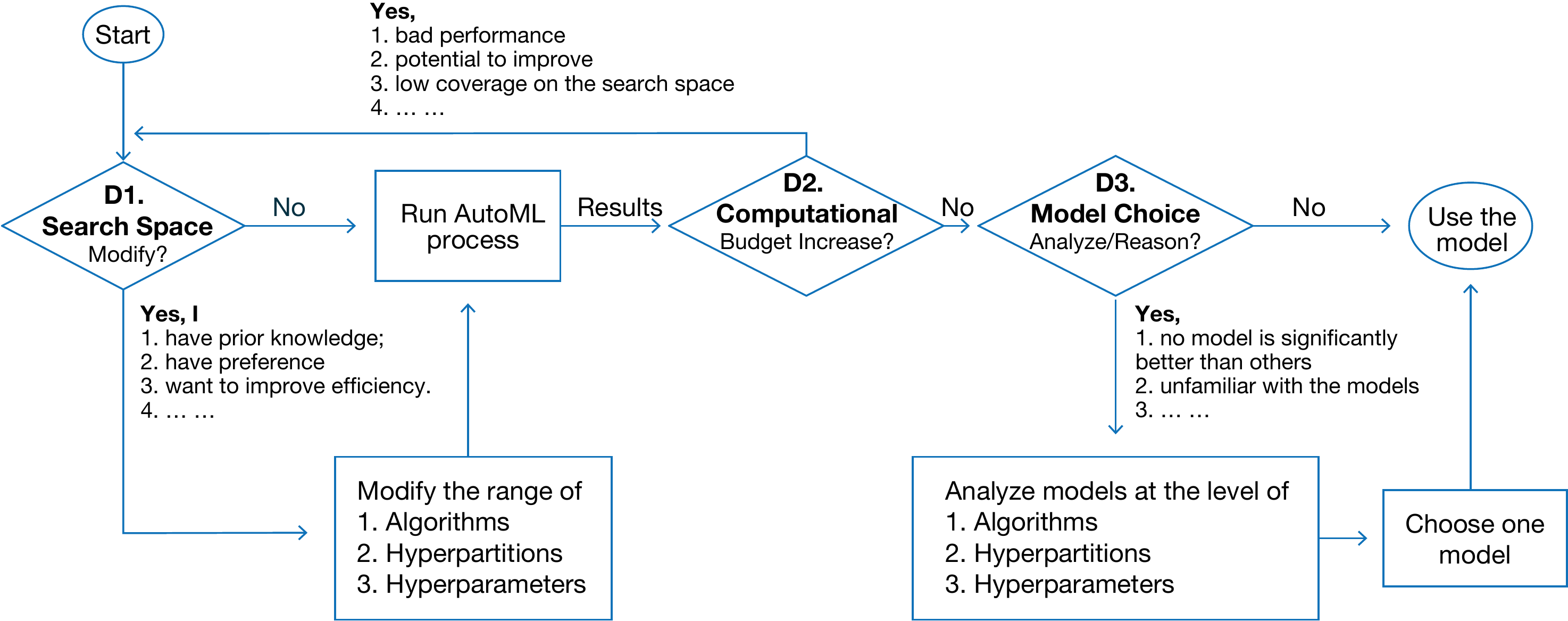}
\caption{A workflow of using AutoML. 
}
\label{fig:workflow}
\end{figure*}

\subsection{Visualizing Machine Learning Models}
\qianwen{
In recent years, there is a trend for combining visualization with machine learning to help people understand, diagnose, improve, and apply machine learning models.
}



\qianwen{
Various visual analytics tools have been developed for opening the black box of different machine learning models, including generative models~\cite{liu2018dgm, wang2018ganviz}, 
reinforcement learning~\cite{zahavy2016grayingDQN}, 
convolutional neural networks~\cite{zeng2017cnncomparator, liu2017cnnvis, liu2018deeptracker}, 
and recurrent neural networks~\cite{ming2017rnnvis, strobelt2018lstmvis}.
These tools provide guidance for model developers to understand, diagnose, and refine machine learning models.
However, in these tools, the requirements of model users are not thoughtfully considered.
}

\qianwen{
To assist in applying machine learning models, some visual analytics tools analyze model behavior on the data instance level without opening up the algorithm black box~\cite{ren2017squares, amershi2015modeltracker, zhang2019manifold}.
For example, Squares~\cite{ren2017squares} reveals the model mistakes at the instance level and connects summary statistics (\eg, accuracy) with individual data instances, thereby helping practitioners analyze model performance.
However, these tools focus on performance analysis and cannot be directly applied to AutoML, in which the configurations (\ie, algorihtm, hyperpartition, hyperparameter) of many searched models need to be analyzed.
}

\section{System Requirements and Design}
\label{sec:design_requirements}


\subsection{Goals \& Target Users}
The main goal of ATMSeer is to help people efficiently search, analyze, and choose machine learning models for their own tasks.
\qianwen{The target users of ATMSeer have a certain level of expertise in machine learning}, 
but previously suffered from a time-consuming and error-prone manual search when developing machine learning models.

\vspace{-1mm}
\subsection{Data Abstraction}


\qianwen{An AutoML process can be regarded as training a sequence of models on a given dataset. At each step, given the performance of previous models, the AutoML algorithm selects a new model to train and evaluate. Each model in an AutoML process can be treated as a multivariate data point with four types of attributes: algorithm (categorical variable), hyperpartition (set of categorical variables), hyperparameter (set of numerical variables), and performance (numerical variable).}

\vspace{-2mm}
\subsection{User Interview}
We conducted semi-structured interviews with six participants to understand how they choose machine learning models and what opportunities exist to improve the experience.
We recruited participants through reaching out to personal contacts.
Three participants were from diverse backgrounds (\ie, biology, urban planning, finance) with experience in developing machine learning models for their domain problems and three participants were machine learning experts.

The interview consisted of three parts and lasted approximately 45 minutes for each participant.
First, the participants were asked to describe their experience in developing machine learning models.
Second, we introduced and discussed AutoML with them, and asked for their expectations of and concerns about AutoML.
Third, the participants were asked to use a pilot system to solve a classification problem and comment on their experiences.
Three participants used their own data and the other three used example data provided by us.
Details of the pilot system are provided in the supplementary material.


\begin{figure*}[!htb]
    \centering
    \vspace{-5mm}
    \includegraphics[width=0.9\linewidth]{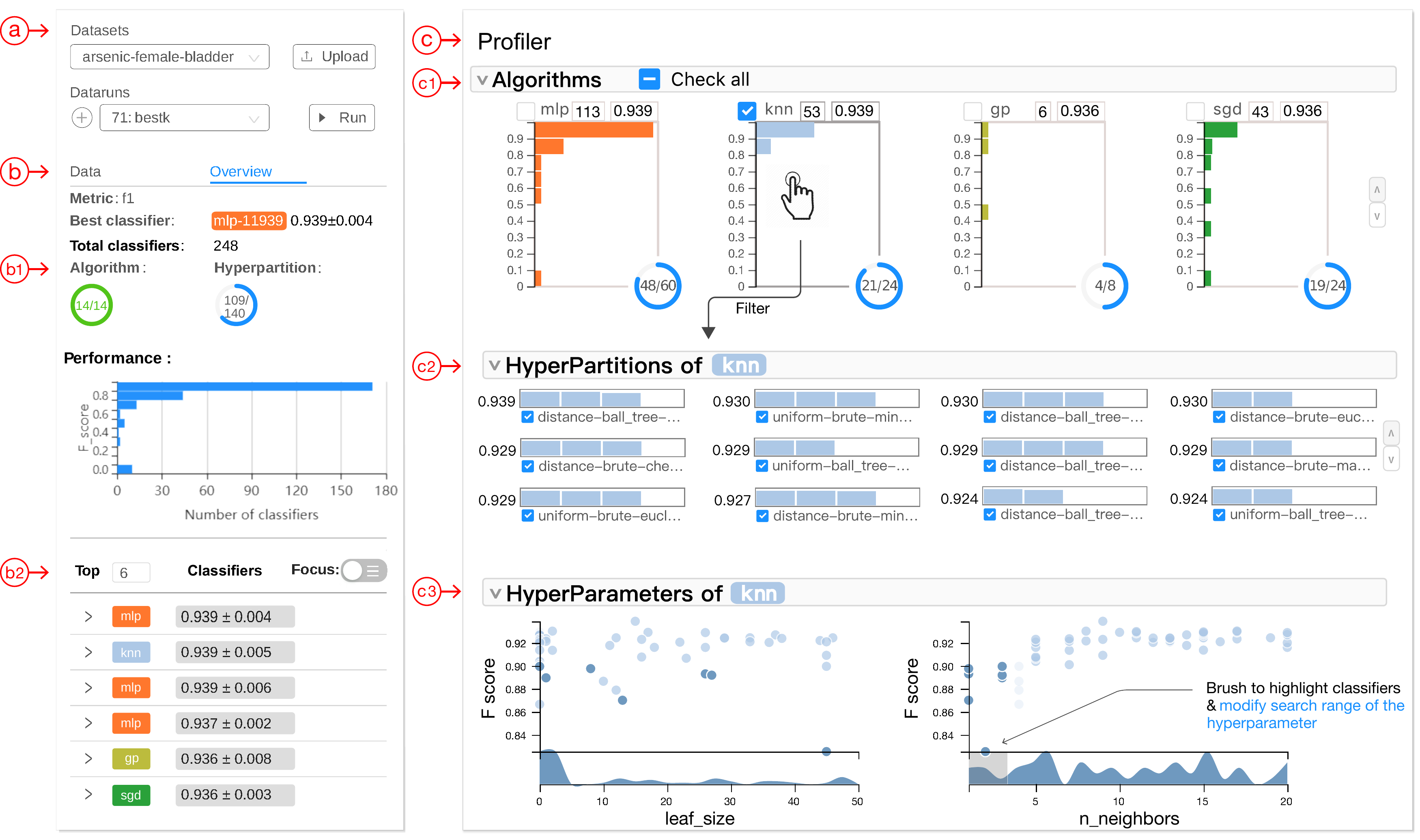}
    \caption{The Interface of ATMSeer. The user creates/resume AutoML process using the control panel (a), observe the high-level statistics of an AutoML process in the overview panel (b), and analyze the process in different granularities with the AutoML profiler (c). 
    }
    \label{fig:interface}
\end{figure*}

\vspace{-2mm}
\subsection{The Workflow}


\qianwen{
Based on the interview, we identify three factors that the participants most care about:
search space (\eg, \textit{``How many algorithms will be searched?''}), 
computational budget (\eg, \textit{``How long will the process run?''}),  
and model choice (\eg, \textit{``Which model is the best among the searched models?''}).  
The three factors correspond to the three key decisions (D1$-$D3) during the use of AutoML and demonstrate the necessity of human involvement.
We connect the three decisions according to how the participants use the pilot system and their current practice of developing models, thereby summarizing a workflow of using AutoML, as shown in \autoref{fig:workflow}.
}

\noindent
\textbf{D1. \qianwen{Modify} Search Space. }
To incorporate human knowledge and improve the search efficiency, AutoML systems usually allow users to configure settings~\cite{auto-sklearn, swearingen2017atm, Olson2016auto}. 
\qianwen{Participants stated that they modified} the search space based on their prior knowledge (\eg, \textit{``the k-nearest neighbors algorithm usually has a good performance on my protein structure dataset and I want to try this algorithm first''}) or their observations of the ongoing search (\eg, \textit{``the random forest algorithm is performing well and is more stable than other algorithms''}).

\noindent
\textbf{D2. \qianwen{Adjust} Computational Budget. }
AutoML tries to find a suitable model by searching through a large set of available machine learning models with limited computational budget (\eg, running time).
There exists a trade-off between the model performance and the computational budget of AutoML.
The participants decided whether to continue an AutoML process based on the performance of the searched models, the potential for subsequent performance improvement, and their acceptable expenses for the AutoML services.

\noindent
\textbf{D3. \qianwen{Reason} Model Choice. }
By default, AutoML returns the model with the best performance score. 
\qianwen{
However, instead of directly using the model with the highest performance score, participants expressed the need to reason the model choice according to the search space (\eg, \textit{``maybe a good algorithm/hyperparameter hasn't been searched''}) or some domain-specific requirements (\eg, \textit{``I prefer models that are robust to the change of hyperparameters''}). 
}

\vspace{-2mm}
\subsection{Design Requirements}
We then distilled the following design requirements to assist in making decisions (D1$-$D3). 

\noindent
\textbf{R1. Offer an overview of the AutoML process.} 
An overview of all searched models can help users learn basic information about the process, such as the number of searched models and how the best performance changes over time
[D2].

\noindent
\textbf{R2. Connect models with the search space.}
Users should be able to analyze models in the context of the search space. 
This enables users to reasone the model choice [D3] and to modify the search space [D1].

\noindent
\textbf{R3. Offer guidance for modification.} 
Guidance should be provided to assist users in modifying the search space [D1].

\noindent
\textbf{R4. Allow in-situ search space configuration.} 
The configuration of search space is usually complex and difficult to memorize. Users should be allowed to switch seamlessly between the observation of the current process and the modification of the search space [D1].

\noindent
\textbf{R5. Support multi-granularity analysis.}
The search space usually has a hierarchical structure (\ie, algorithms, hyperpartitions, hyperparameters).
A multi-granularity analysis of searched models should be supported
to help users monitor and analyze the searched models [D3].

\begin{figure}[!htb]
	\centering
	\includegraphics[width=1.0\linewidth]{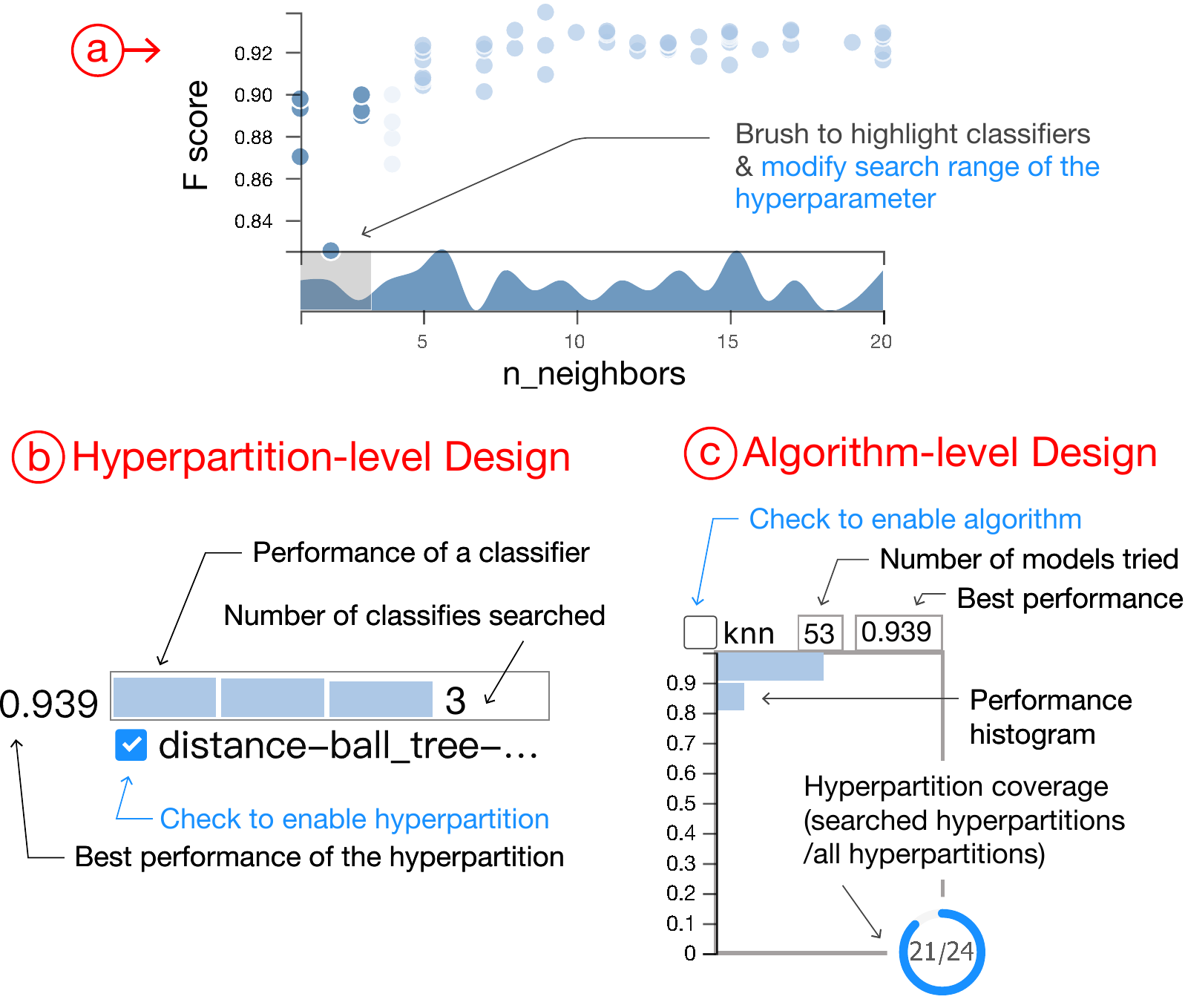}
	\caption{Detail of ATMSeer interface elements in the (a) hyperparameter-level, (b) hyperpartition-level, and (c) algorithm-level views.}
	\vspace{-5mm}
	\label{fig:interface_detail}
\end{figure}

   

\begin{figure}[!htb]
    \centering
    \includegraphics[width=0.9\columnwidth]{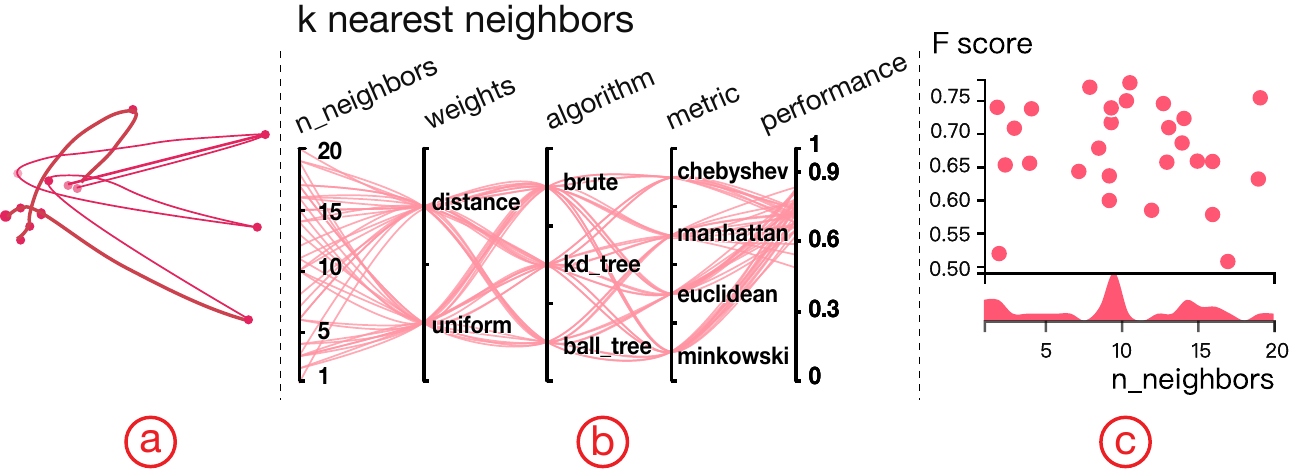}
    \caption{Alternative designs of the hyperparameter-view: a PCA mapping (a), parallel coordinates (b), and scatter plots (c). }
    \vspace{-5mm}
    \label{fig:alternative_hyperparameter}
\end{figure}
\vspace{-2mm}

\section{ATMSeer}
This section describes the design and implementation choices of ATMSeer, \qianwen{an interactive visualization tool that enables users to refine the search space of AutoML and analyze automatically-generated results.}
\subsection{System Overview}
ATMSeer is implemented as a client-server application. 
The server coordinates three components: AutoML, data storage and model storage. 
The server provides the client with various APIs to create, configure, start, and pause an AutoML process, and to summarize the recorded data in AutoML processes. 
As the client, the visual interface provides graphical controls for AutoML processes and maps the summary data to visualization. 
\subsection{Interface}
The interface of ATMSeer consists of three parts (\autoref{fig:interface}):
\begin{itemize}
    \item the control panel (a), which allows users to upload a new dataset or select an existing dataset and create or resume an AutoML process.
    \item the overview panel (b), which shows high-level statistics of the dataset and the AutoML process.
    \item the three-level AutoML Profiler (c) for analyzing the AutoML process at different granularities.
\end{itemize}
\vspace{-2mm}

\subsubsection{AutoML Overview}
The overview panel (\autoref{fig:interface}(b)) summarizes high-level information (\textbf{R1}) of an AutoML process in two aspects: general summary (b1) and top models (b2). 
In addition to the best performance score and the total number of models, two coverage metrics show the percentage of searched algorithms and hyperpartitions. Next, the performance distribution summarizes the performance of all tried models in a histogram. The top $k$ models are listed for the users to compare and choose. 
Users can focus their analysis on the top models by enabling the \textit{``focus mode''}, which highlights the corresponding algorithms and hyperpartitions in the detail views.

\vspace{-2mm}
\subsubsection{AutoML Profiler}
After an AutoML process exhausts its budget, the user must carefully decide whether to resume the process with an increased budget (\textbf{D2}) and/or a refined search space (\textbf{D1}).
It can also be challenging for users to decide which model to choose and if the chosen model is really the best (\textbf{D3}). 
How well does each type of algorithm perform? How many models have been tried for each algorithm? Can AutoML find better models with an increased budget and better configuration? Are some highly-ranked models likely to achieve better generalization performance than others?
The AutoML Profiler summarizes an AutoML process at three levels of granularity to help users answer these questions. From macro to micro, these include algorithm-level, hyperpartition-level, and hyperparameter-level. The latter two levels, which are only shown on demand, are designed for advanced users who need to evaluate or configure an AutoML process at a finer granularity.


\textbf{Algorithm-level View} (\autoref{fig:interface}(c1)) visualizes the performance distribution of each machine learning algorithm as a histogram. As shown in (\autoref{fig:interface_detail}(c)), important statistics of the algorithms, such as best performance and hyperpartition coverage, are displayed with each histogram. 
The algorithms are sorted according to their best performance in descending order. 
The algorithm-level view enables users to compare different algorithms (\textbf{R5}) with respect to performance distribution and the number of tried models. Users can evaluate the robustness and performance of each algorithm and gain intuition for modification (\textbf{R3}). 

\textbf{Hyperpartition-level View} (\autoref{fig:interface}(c2)) summarizes different hyperpartitions of selected algorithms. 
A hyperpartition is a configuration of an algorithm with fixed non-tunable hyperparameters (only numeric hyperparameters can be tuned). Different hyperpartitions of an algorithm can have very different properties (\eg, SVM with linear kernel vs. polynomial kernel). 
The hyperpartition-level view is designed to help \qianwen{advanced} users to analyze the search space (\textbf{R2}) and compare  hyperpartitions (\textbf{R5}). For a selected algorithm, its hyperpartitions are visualized as a list of progress bars. Models are pushed into their corresponding progress bars according to their hyperpartitions as small boxes, whose height denotes the performance of the model (\autoref{fig:interface_detail}(b)).  

\textbf{Hyperparameter-level View} (\autoref{fig:interface}(c3)) shows the relation between performance and hyperparameters of a selected algorithm. 
For each tunable hyperparameter, a scatter plot is presented to compare hyperparameter values and performance scores. Each model is visualized as a point in each of the scatter plots. We also include an area plot showing the distribution of the hyperparameter below each scatter plot to help users evaluate the coverage of the hyperparameter space \qianwen{(\ie, which values have been extensively tried and which have not)} (\textbf{R2}).
It also helps users learn how each hyperparameter influences the performance: at what values a model gets a generally good performance and where it does not. 
This information can be used as important hints for improving the configuration of the search space (\textbf{R3}).

During the development of ATMSeer, we experimented with two common multivariate visualization techniques as design alternatives (\autoref{fig:alternative_hyperparameter}): principal component analysis (PCA) mapping and parallel coordinates. 
We gathered preliminary feedback from three target users that we interviewed (\autoref{sec:design_requirements}).
The PCA mapping was rejected by all users, since it loses the detail of hyperparameter values, which are important for analyzing and comparing models.
Parallel coordinates visualize each hyperparameter in the high-dimensional coordinate space as a vertical line and were well-accepted by the users. 
However, we found that users needed to perform intensive interactions with the parallel coordinates during the analysis. 
We also noticed that most users are only interested in investigating the relationship between performance and a single hyperparameter at a time, possibly resulting from the fact that high-dimensional relationships are perceptually challenging for humans.
Compared with parallel coordinates, multiple scatter plots are intuitive, simple, and preferred by all users.
As a result, we adopt scatter plots as the final design for the hyperparameter-level view.

\vspace{-1mm}

\subsection{Interaction Design}
\subsubsection{Real-time Control} 
The ATMSeer interface is updated dynamically, which allows users to monitor and analyze AutoML processes in real-time. The users can also perform a ``run-reconfigure-run'' workflow -- they can pause and reconfigure an AutoML process and then restart it from its previous state.

\subsubsection{In-Situ Search Space Configuration} 
A well-chosen search space could improve the search efficiency of AutoML. Configuring the search space also helps integrate a user's prior knowledge into the AutoML algorithm.
However, configuring the search space can be a challenging task due to its complex hierarchical structure (\ie, different algorithms, categorical and numeric hyperparameters). 
As shown in \autoref{fig:interface_detail}(a), ATMSeer provides an in-situ configuration which embeds in the three-level Profiler. It allows users to easily modify the search space at the same place they observe and analyze the search models (\textbf{R4}). 
\begin{figure}
    \centering
    \includegraphics[width=\columnwidth]{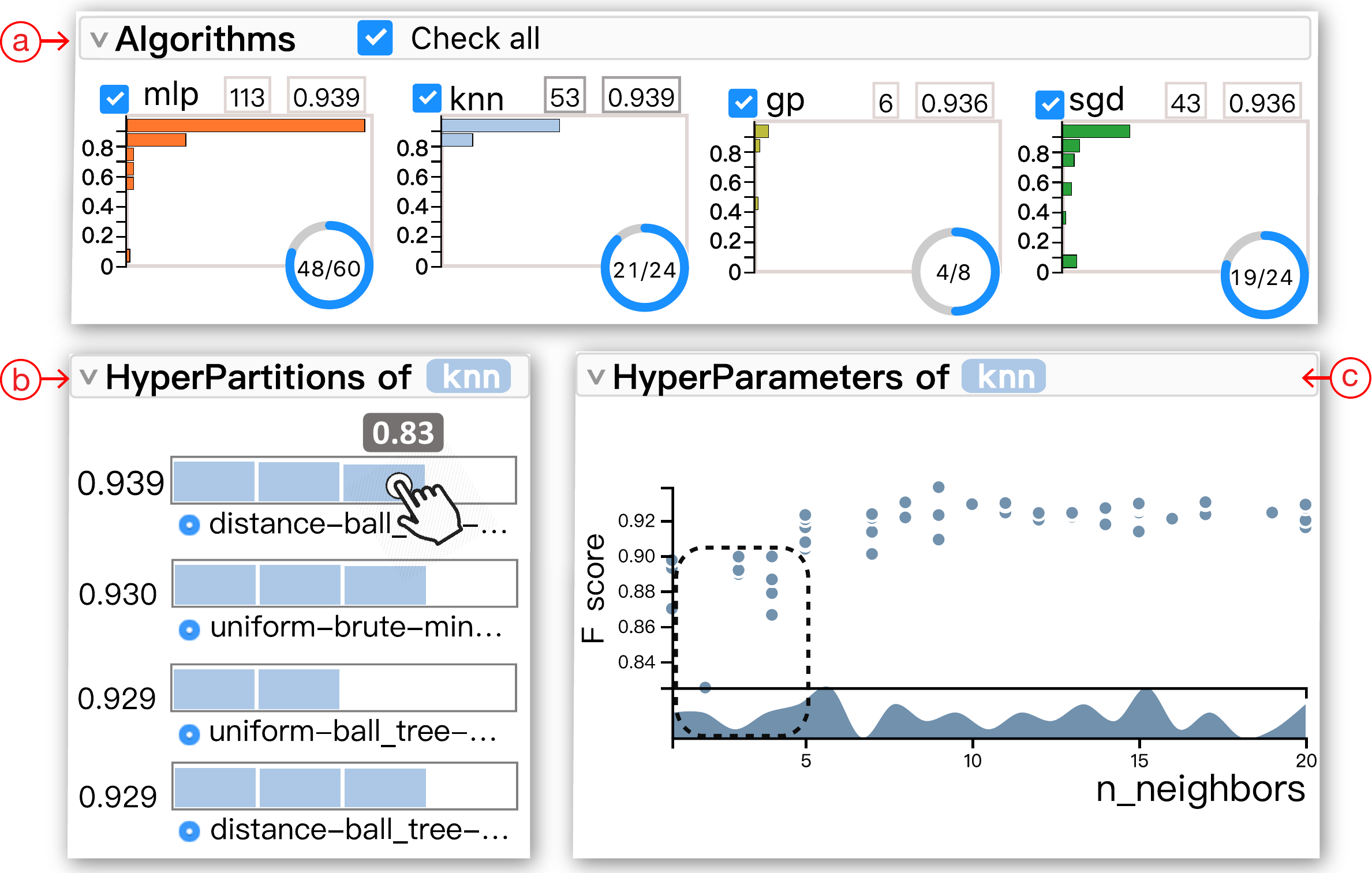}
    \caption{The use case of ATMSeer to select and understand models. (a): The top 4 algorithms have similar best performances but KNN has a more stable performance distribution. (b): There is no obvious relationship between the choice of KNN hyperpartition and the model performance. (c): A small value of \texttt{n\_neighbors} leads to low performance.}
    \vspace{-3mm}
    \label{fig:case_select}
\end{figure}

\section{Case Study}
\label{sec:case_study}
The case studies are conducted in collaboration with two ML experts (denoted as E1 and E2) that we interviewed in \autoref{sec:design_requirements}.
We use F1-score with 10-fold cross validation as the performance metric.
The machine learning algorithms used in the case studies include support vector machine (SVM), extra-trees (ET), linear models with SGD training (SGD), $k$-nearest neighbors (KNN), random forest (RF), multi-layer perceptron (MLP), and Gaussian process (GP).

\subsection{Select and analyze models (D2, D3)}
In this case, we illustrate how ATMSeer helps users select and analyze the searched models.
E1 wants to find a model for the \texttt{arsenic-female-bladder} dataset~\cite{arsenicDataset} using ATMSeer.
This dataset classifies 559 female patient records as positive (bladder cancer) or negative (healthy).
ATMSeer first searched 250 models for this dataset.
Observing that the best performance score is 0.939 and the algorithm coverage is 100\% in the overview panel (\autoref{fig:interface}(b2)), E1 is satisfied with the AutoML results.
E1 then decides to stop the search (\textbf{D2}) and to choose a model from the already searched models.

E1 first examines the top 10 models in the leaderboard.
The top 10 models have similar performance scores (from 0.936 to 0.939) and belong to four different algorithms (\ie, MLP, KNN, GP, SGD).
Since these models have similar performance scores, E1 thinks it would be better to choose a model by comparing the characteristics of the algorithms.
E1 then compares these four algorithms in the algorithm-level visualization.
He finds that the performance distribution of KNN is concentrated on the top, implying that KNNs have generally good performance on this dataset. He decides that KNN should be a good choice (\textbf{D3}).

E1 also wants to learn why some KNNs have less satisfying performance (\ie, F1-score $<$ 0.9) to increase his confidence in using the model.
He clicks KNN in the algorithm-level view to reveal more information in the hyperpartition-level view (\autoref{fig:case_select}(b)).
He notices that the difference between the best performance of each hyperpartition is not significant.
Meanwhile, one hyperpartition can have both strongly performing models and poorly performing models (\eg, a model in the first hyperpartition has a performance of 0.83).
E1 then clicks the hyperparameter-level view to observe more detailed information and finds that the choice of hyperparameter directly influences the performance.
As shown in \autoref{fig:case_select}(c), most poorly-performing models have a small ``number of neighbors''.

\begin{figure}
    \centering
    \includegraphics[width=1.0\columnwidth]{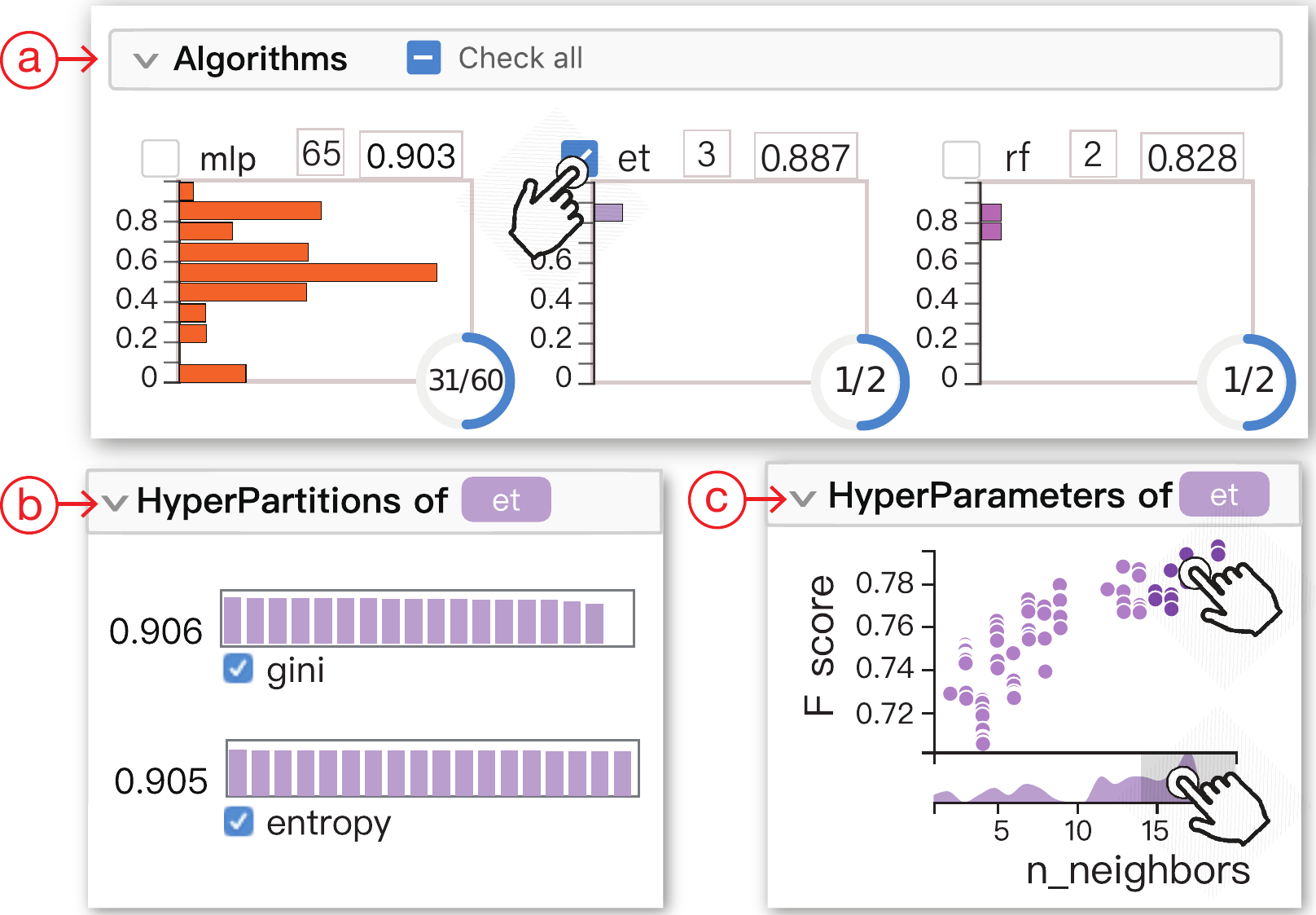}
    \caption{Using ATMSeer to modify the search space of an AutoML process. (a): Only ET algorithm is selected. (b): The best performance of the two ET hyperpartitions increases to 0.906 and 0.905. (c): The range of the \texttt{max\_feature} is set to $[0.7, 1.0]$ and the best performance increases to 0.922.}
    \vspace{-4mm}
    \label{fig:case_refine}
\end{figure}

\begin{figure}
    \centering
    \includegraphics[width=\columnwidth]{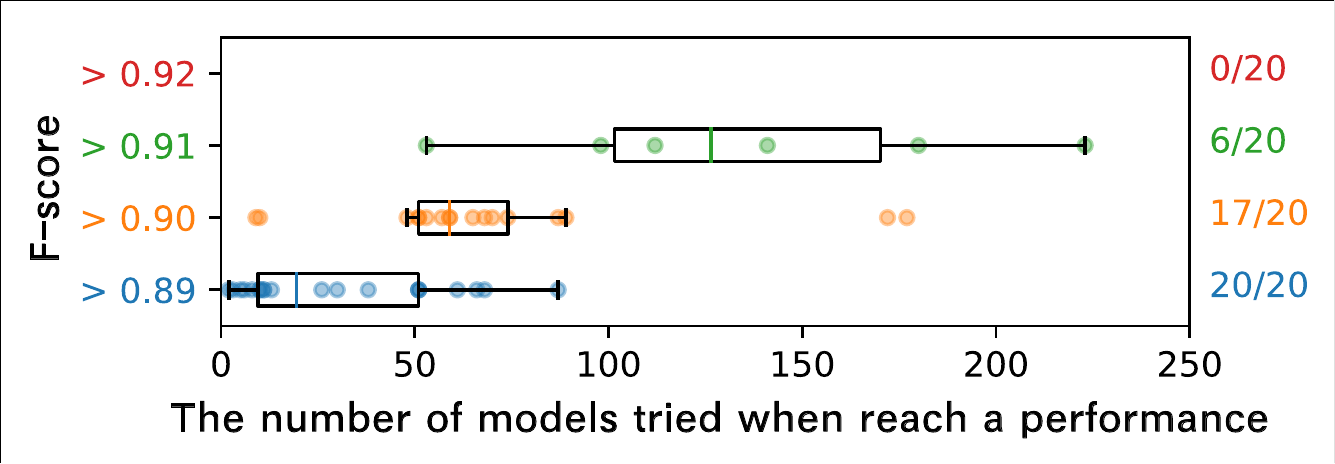}
    \caption{The performance of 20 AutoML processes. X-axis represents the number of models tried when a process first reaches a certain performances. Among 20 processes, all 20 processes achieves a best performance of over 0.89, 17 over 0.90, 6 over 0.91, and none over 0.92. }    
    \vspace{-5mm}
    \label{fig:case_performace_distribution}
\end{figure}
\vspace{-2.0mm}

\subsection{Refine Search Space (D1, D2)}
Next, we illustrate how ATMSeer helps users modify an AutoML process and improve the performance.
E2 wants to find a model for the Friedman Dataset \texttt{fri\_c3\_1000\_10}~\cite{friedman2002stochastic}, a synthetic binary classification problem with 1000 instances and 10 features.
With strong domain knowledge, E2 wants to have more control over the AutoML process.
He first tries all algorithms to analyze which one is better for this dataset.
After searching 150 models, the algorithm coverage reaches 100\% and he suspends the process.
After observing the algorithm-level view (\autoref{fig:case_refine}(a)), E2 expresses a preference for the second best algorithm, ET.
ET is tried for only three times and its performances is comparable to the best algorithm, MLP. 
In addition,
ET has a concentrated performance distribution between 0.8 and 0.9.
Thus, E2 modifies the algorithm-level configuration to focus on searching ET to see if further improvements could be achieved.
After searching another 30 models, the best performance increases from 0.887 to 0.906.
Opening the hyperpartition-level view (\autoref{fig:case_refine}(b)), 
E2 finds the performance distributions of ET-Entropy and ET-Gini are similar. 
This is consistent with his prior knowledge that ETs with Gini or Entropy measure have similar performance in general~\cite{raileanu2004theoretical}.

E2 then checks the hyperparameter-level view (\autoref{fig:case_refine}(c)) and finds that the value of \texttt{max\_features}\footnote{In ET, \texttt{max\_features} is a node splitting criterion given by the ratio of the size of random feature subsets to the total number of features.} directly influences the performance.
Based on his observation of the hyperparameter-level view, he concludes that choosing \texttt{max\_} \texttt{features} between 0.7 to 1 leads to higher performance, which, however, conflicts with his experience:
\textit{``The empirical good value of \texttt{max\_} \texttt{features} is around $\frac{\sqrt{n}}{n}$ [$n$ is the number of features] \qianwen{~\cite{louppe2014understanding}}, which means I would set it to around $0.3$ for this dataset.''}
E2 comments that
\textit{``this makes sense since empirical values are usually not optimal.''}
E2 then modifies the range of \texttt{max\_features} to $[0.7, 1.0]$ (\autoref{fig:case_refine}(c)) and searches another 50 models. The best performance score increases to $0.922$.
Since no further performance improvement occurs in the last 20 searched models, E2 stops the search process and chooses the best ET model.

Interested in assessing how the involvement of humans improves an AutoML process, we run 20 AutoML processes without human interference for comparison.
We let each AutoML process independently search 250 models.
As shown in \autoref{fig:case_performace_distribution}, among 20 processes, 17 reach a performance score of over $0.90$ and 6 reach $0.91$, but none reaches $0.92$. 
In comparison, the expert achieves the best performance of 0.922 within the 210th model, which shows that human involvement has the potential of improving both the performance and the efficiency of AutoML.
This improvement might be caused by the fact that the human could modify the search space in a more aggressive way (\ie, only choose one algorithm and reduce the range of a hyperparameter to 30\% of its original range).

\section{Expert Interview}
To evaluate ATMSeer, we conduct interviews with two closely collaborating experts (E1 and E2, the same experts we collaborated with in \autoref{sec:case_study}) with particular expertise in AutoML.
E1 is a co-author of an AutoML framework ~\cite{swearingen2017atm}.
We collected their feedback about ATMSeer through weekly meetings over more than two months.
Based the discussion, we summarize three main usage scenarios of ATMSeer.



\vspace{-2mm}
\subsection{Knowledge Distillation from AutoML}
\begin{figure}
    \centering
    \includegraphics[width=\columnwidth]{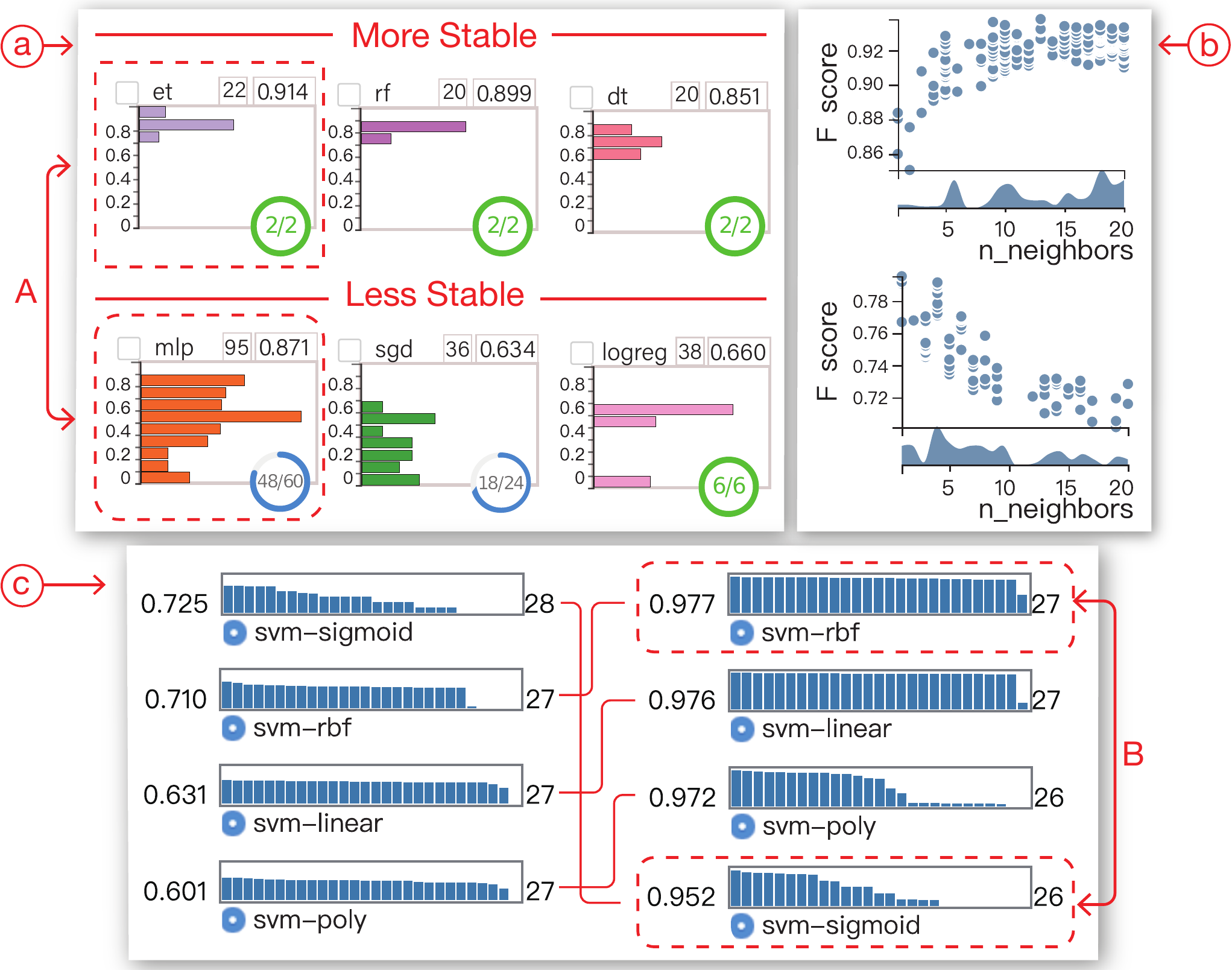}
    \caption{(a): Some algorithms  have more stable performance distributions than other algorithms (using Friedman Dataset fri\_c3\_500\_10~\cite{friedman2002stochastic}). (b): The suitable range of n\_neighbors in KNN varies from dataset to dataset (top: Quake Dataset~\cite{simonoff2012smoothing}; bottom: Machine CPU Dataset from UIC\cite{UIC2017}). (c): The suitable SVM hyperpartition for different datasets varies (top: Arsenic Female Bladder Dataset~\cite{arsenicDataset}, bottom: Ringnorm Dataset~\cite{ringnorm}).}
    \vspace{-5mm}
    \label{fig:expert_interview}
\end{figure}

ATMSeer can help people better understand and apply machine learning algorithms.
AutoML enables quick experimentation with a large number of models, whose results could provide useful knowledge to ML researchers and practitioners.
As shown in \autoref{fig:expert_interview}(a), ATMSeer shows that some algorithms (\eg, ET, RF, KNN) tend to have a stable performance distribution while other algorithms (\eg, SGD, MLP, SVM) are more prone to generating poorly-performing models.
For the same algorithm, the strongly-performing hyperpartitions (\autoref{fig:expert_interview}(c)) and hyperparameters (\autoref{fig:expert_interview}(b)) vary from dataset to dataset. 
These findings can inform users of the importance of hyperparameter tuning for certain algorithms.

E2 commented that being able to match prior knowledge about machine learning to the visualizations produced by ATMSeer creates confidence in the underlying AutoML process and increases the likelihood of adopting AutoML.
He also believed that ATMSeer could function as an educational tool for machine learning, which helps people better understand the behavior of unfamiliar algorithms.
\vspace{-2mm}
\subsection{Human-Machine Interaction in AutoML}
Both experts appreciated the human-machine interaction introduced in ATMSeer.
They believed such interaction can improve an AutoML process and enhance user experience. 
They commented that \textit{``human observation and prior knowledge sometimes can be more efficient than AutoML algorithms, especially when there is a large search space and limited computational budget.''} 
E1 said that \textit{``users with more domain knowledge, such as myself, are usually critical of automated methods and like to be in control. I don't like getting a score back and hearing `trust me.' ''}

\subsection{Diagnosis of AutoML}
Even though ATMSeer is initially developed for AutoML end users, 
our expert interviews suggest that ATMSeer can also help AutoML developers diagnose AutoML algorithms.

For example, using ATMSeer, the experts identified that the AutoML method proposed by \citeauthor{swearingen2017atm}~\cite{swearingen2017atm} seems to be biased in favor of certain algorithms.
As shown in \autoref{fig:expert_interview}(a), 
MLP was searched many more times than ET even though their best performance was similar and the mode of the performance distribution of MLP was much lower than that of ET.
E1 quickly identified the reason for this phenomenon using ATMSeer.
In the ATM framework, each hyperpartition is modeled as separate arm in a multi-armed bandit framework and competes with others for the opportunity to be searched.
In this case, algorithms with a small number of hyperpartitions (\eg, RF, ET) tend to be searched less, motivating a hierarchical extension.

Another issue the experts identified was that the search frequencies of strongly-performing and poorly-performing hyperpartitions were sometimes similar.
As shown in \autoref{fig:expert_interview}(b), SVM with RBF kernel, having a higher best performance score and a more stable performance distribution, is likely better for the given dataset compared with SVM with Sigmoid kernel.
However, SVM-RBF wasn't tried notably more times than SVM-Sigmoid (27 vs. 26).
E2 thought this reflected an inappropriate reward function in the underlying AutoML algorithm: the search frequency of a hyperpartition (\ie, the reward) was determined by the average performance of its best $k$ models ($k=5$ in this case).
The search frequencies of SVM-RBF and SVM-Sigmoid were similar because their best $k$ models were similar.
E2 thought this made him reconsider the design of the reward function.

\begin{table*}
    \centering
    \vspace{-5mm}
    \begin{tabular}{ c | p{4cm} | p{5cm}| p{5.5cm}  }
    \hline
         & Algorithm & Hyperpartition &  Hyperparameter  \\
    \hline
       Clicks  & \textbf{51\% (13/13)} & \textbf{28\% (10/13)} & \textbf{21\% (7/13)} \\
    \hline
      \multirow{3}{4em}{Questions}  & Q1.find the best performed algorithm: \textbf{13/13} 
                                    & Q4.find the algorithm with most hyperpartitions: \textbf{13/13} 
                                    & Q7.find the suitable range of ``n neighbors'' in KNN: \textbf{11/13} \\
                                    \cline{2-4}
                                    & Q2.find the most stable algorithm: \textbf{13/13}  
                                    & Q5.find the worst performed hyperpartition in MLP: \textbf{12/13}   
                                    & Q8.find the hyperparameter that influences the performance of SGD: \textbf{11/13}  \\
                                    \cline{2-4}
                                    & Q3.find the most frequently searched algorithm: \textbf{13/13} 
                                    & Q6.find the best performed hyperpartition in SVM: \textbf{13/13} 
                                    & \\
    \hline
    \end{tabular}
    \caption{Results on the interaction with and the understanding of three levels of information.}
    \vspace{-5mm}
    \label{tab:three_level}
\end{table*}

\section{User Study}
The visual analysis of AutoML processes is a relatively new problem. To the best of our knowledge, there are no similar tools for comparison.
\qianwen{
At the same time, we found that comparison with a baseline system\footnote{https://github.com/HDI-Project/ATMSeer/tree/dev-zhihua-baseline} was unfair. Since the baseline system failed to provide detailed information about the AutoML results, users are unable to make informative decisions and modify the search process.
Therefore, instead of an unfair benchmarking, we believe it more interesting and important to investigate user behavior under the characterized workflow with ATMSeer.
}



\subsection{Participants and Apparatus}
We recruited 13 graduate students by email (11 male and
2 female, age 22--30, $\mu = 24.46$, $\sigma = 2.66$), denoted as P1-P13.
All participants had experience in machine learning or data science but none of them had prior experience with AutoML.
Each participant was rewarded with a gift card worth \$10. 
All studies were conducted with a 24-inch 1920 $\times$ 1200 monitor, a mouse, and a keyboard.



\subsection{Datasets \& Tasks}
The participants were asked to perform two tasks that mimic the workflow described in~\autoref{sec:design_requirements}.

\begin{itemize}
    \item \textbf{T1:} try their best to find a model with good performance for an given dataset using AutoML. 
    \item \textbf{T2:} analyze a given AutoML process and answer 13 questions related to \textbf{D1}-\textbf{D3}.
\end{itemize}

For \textbf{T1}, we use the German Credit Dataset~\cite{UIC2017}, which classifies 1000 loan applicants as good or bad credit risks based on 20 features.
For \textbf{T2}, to ensure a fair comparison across participants, we pre-run an AutoML process for 200 models on the artificially-generated Friedman \texttt{fri\_c3\_500\_10} dataset~\cite{friedman2002stochastic}.
\qianwen{
In our 13 question survey, \textbf{Q1-Q8} are objective questions that investigate users' understanding of the three-level profiler (\autoref{tab:three_level}); \textbf{Q9-Q13} are subjective questions that investigate the information users refer to when making decisions about increasing the computational budget, modifying the search space of algorithms, modifying the search space of hyperpartitions, modifying the search space of hyperparameters, and choosing a model, respectively.
}


\subsection{Procedure}
The study began with a tutorial session, in which the tasks and the usage of ATMSeer were introduced to the participants.
When performing the tasks, participants were free to ask questions and were encouraged to think-aloud.
We deemed \textbf{T1} as complete when the participant was satisfied with the AutoML results and chose a model for the given dataset.
In \textbf{T2}, users were allowed to skip questions that they did not know.
The click activities of participants were automatically recorded.
Finally, participants completed four usability questions using a 5-point Likert scale (1 for strongly disagree and 5 for strongly agree).
A post-study interview was conducted to collect more detailed feedback from the participants.
Each user study session lasted about 40 minutes.



\begin{figure}[b]
    \centering
    \includegraphics[width=0.8\columnwidth]{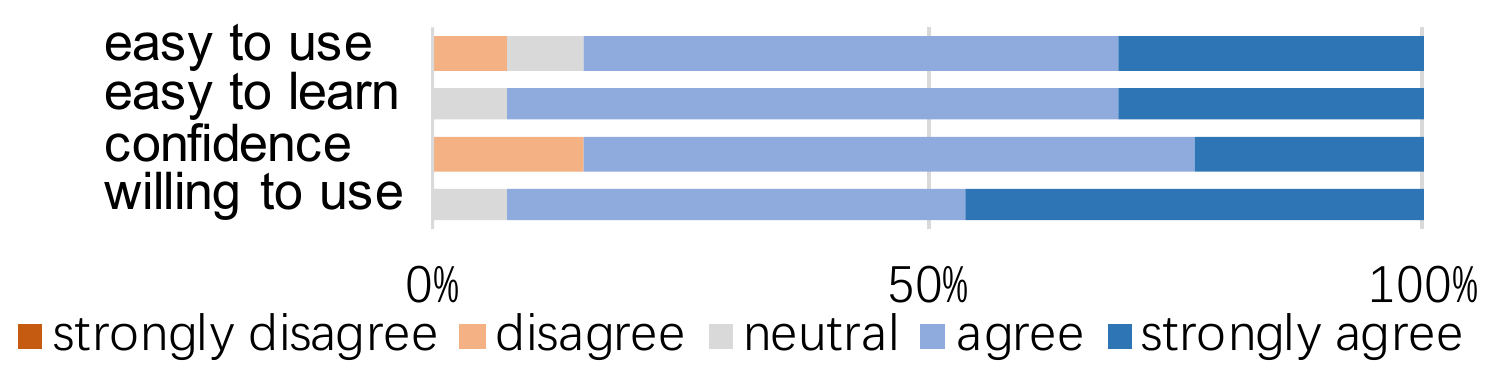}
    \caption{Results of the usability questionnaire.}
    \label{fig:usability}
\end{figure}


\subsection{\qianwen{Usability}}
The result of the usability questionnaire is summarized in \autoref{fig:usability}. Overall, most participants agreed that ATMSeer was easy to learn and easy to use.
Most participants (strongly) agreed that they were confident in their selected model (84.6\%) and were willing to use ATMSeer in the future (92.3\%).
P11 disagreed that ATMSeer was easy to use, commenting that 
\textit{``I cannot remember the meaning of every hyperparameter and am not familiar with every algorithm.''}
P11 and P13 disagreed that they were confident in the selected model and desired additional validation using their familiar tools.

For the objective questions (Q1$-$Q8) in \textbf{T2},
participants answered fluently and correctly most of the time (99/104). 
This indicated the usability of ATMSeer in enabling users to analyze the searched models.
Among the five errors/missing responses, one was a careless mistake (the participant mistook SVM as SGD).
The other four came from the hyperparameter-level questions, which we found to have been caused by a low familiarity with the ML models, according to the post-study interview.
\qianwen{For example, two participants skipped Q8 because that they were unfamiliar with SGD classifiers and were not confident about the correctness of their observations. }




\subsection{Revisiting the Workflow}
The analysis of user behavior helps us to reflect on the workflow in ~\autoref{sec:design_requirements}.

\noindent
\textbf{D1: Modify Search Space. } 
Among 13 participants, 10 follow a coarse-to-fine strategy to refine the search space. 
Specifically, after searching some models, participants first refined the search space at a coarse level (\eg, choose 2 to 3 algorithms) based on their observation and continued the search. 
After searching more models, participants then refined the search space at a finer level (\eg, modified the range of a hyperparameter).
\qianwen{
Participants expressed different preferences for refining the different levels of the search space and were less interested in fine modifications (\autoref{tab:three_level}).
}

\noindent
\textbf{D2: Adjust Computational Budget.} 
According to the interview, whether to continue the search with an increased budget is a decision made based on multiple reasons, 
including the \textit{performance histogram} (considered by 9 participants), 
the \textit{number of searched models} (by 5),
the \textit{algorithm coverage} (by 9),
and the \textit{best performance} (by 5).


We also found that people with strong domain knowledge (\ie, self-reported as expert or advanced) were more willing to increase the computational budget so that they could further modify the search space.
Six advanced participants searched on average 117 models ($\sigma=62.24$) while seven novice participants searched 97.5 models ($\sigma=62.49$).

\noindent
\textbf{D3: Reason Model Choice.} 
While all participants preferred the models with good performance, nine participants also expressed preferences for familiar models.
Three participants commented that ATMSeer helped them understand machine learning models, thereby improving their familiarity with a model and their willingness to use it. 


\section{Discussion \& Future Work}

\subsection{General Applicability}
ATMSeer is initially designed for machine learning experts. 
However, based on our expert interviews and user studies, we identify other potential usage scenarios of ATMSeer, including learning machine learning models and debugging AutoML algorithms.
For beginners in machine learning, ATMSeer enables them to observe how the choice of algorithms, hyperpartitions, and hyperparameters influences model performance. 
For AutoML designers, ATMSeer enables them to analyze the results of an AutoML process and identify possible bugs and opportunities for improvement.
We will conduct further investigation in future work.

\subsection{Limited Evaluation}
One limitation of this work is that the user studies are only conducted with 13 participants --- predominantly graduate students.
Further investigation would help validate whether the insights and results can be applied more generally.
Nevertheless, we are encouraged by the fact that ATMSeer was appreciated by these users and got positive feedback.


\vspace{-2mm}
\subsection{Scalability of the Visualization}
Existing AutoML systems support 8--15 algorithms in general~\cite{MLJar, auto-sklearn}. A typical AutoML process would search 100 to 400 models in total~\cite{swearingen2017atm, MLJar, auto-sklearn}. 
ATMSeer uses a categorical color scheme to encode different machine learning algorithms, which most users could distinguish. The visualization can fluently support the analysis an AutoML process with over 1000 models. 
Thus, the effectiveness of ATMSeer is guaranteed for most real-world machine learning tasks.
\vspace{-1mm}
\subsection{Future Work}
We envision improving ATMSeer in several directions. 
First, we plan to further validate ATMSeer in real-world applications with a larger and more diverse group of users. 
ATMSeer is open-sourced.
We will further evaluate it and improve it based on the future feedback.
Second, we intend to support intelligent control of AutoML processes.
One possible direction is to combine ATMSeer with human-in-the-loop reinforcement learning~\cite{mandel2017add} and automatically detect the critical points (\eg, stuck in a local optimum) in an AutoML process where human involvement is needed.
\section{Conclusion}

\qianwen{
In this work, we presented ATMSeer, an interactive visualization tool that supports machine learning experts in analyzing the automatic results and in refining the search space of AutoML.
A workflow of using AutoML was proposed based on the interview with machine learning experts. 
Three key decisions in this workflow --- updating the search space, modifying the computational budget, and reasoning the model choice --- were identified to guide the design and implementation of ATMSeer.
We next proposed a multi-granularity visualization with in-situ configuration to enable users to examine an AutoML process in real time at the algorithm level, hyperpartition level, and hyperparameter level.
A series of evaluations demonstrated the utility and usability of ATMSeer.
The user study suggested that users followed a coarse-to-fine strategy when using ATMSeer and that users with a higher level of expertise in machine learning were more willing to interact with ATMSeer.
}

\begin{acks}
The authors would like to thank all the participants involved in the studies and the reviewers for their constructive comments and valuable suggestions. 
This work is supported by The Hong Kong Bank Foundation under Grant No.HKBF17RG0.

\end{acks}

\end{spacing}

\clearpage

\balance
\bibliographystyle{ACM-Reference-Format}
\bibliography{ref}

\end{document}